\documentclass[journal]{IEEEtran}

\usepackage{times}
\usepackage{epsfig}
\usepackage{graphicx}
\usepackage{amsmath}
\usepackage{amssymb}
\usepackage{algorithm}
\usepackage{algpseudocode}
\usepackage{float}
\usepackage{booktabs}
\usepackage{multirow}
\usepackage{diagbox}
\usepackage{array}
\usepackage{arydshln}
\usepackage{color}
\usepackage{caption}
\usepackage[export]{adjustbox}

\newcolumntype{x}[1]{>{\centering\arraybackslash\hspace{0pt}}p{#1}}
\newcolumntype{z}[1]{>{\arraybackslash\hspace{0pt}}p{#1}}

\usepackage[pagebackref=true,breaklinks=true,colorlinks,bookmarks=false]{hyperref}

\newcommand{\ve}[1]{\mathbf{#1}}
\newcommand{\vv}[1]{\mbox{\boldmath $#1$}}

\ifCLASSINFOpdf
\else
\fi
\begin{document}
%
\title{Preserving Earlier Knowledge in Continual Learning \\ with the Help of All Previous Feature Extractors}
%
%
%


\author{Zhuoyun~Li, 
        Changhong~Zhong, 
        Sijia~Liu, 
        Ruixuan~Wang,
        and~Wei-Shi~Zheng,
\thanks{Z. Li, C. Zhong, S. Liu, R. Wang, and WS. Zheng are with the School of Computer Science and Engineering, Sun Yat-sen University, and also with the Key Laboratory of Machine Intelligence and Advanced Computing, MOE, Guangzhou, 510000, China; \\
Correspondence e-mail: wangruix5@mail.sysu.edu.cn}
}

%
%

\markboth{Manuscript submitted to journal}{}
%



\maketitle

\begin{abstract}
Continual learning of new knowledge over time is one desirable capability for intelligent systems to recognize more and more classes of objects. Without or with very limited amount of old data stored, an intelligent system often catastrophically forgets previously learned old knowledge when learning new knowledge. Recently, various approaches have been proposed to alleviate the catastrophic forgetting issue. However, old knowledge learned earlier is commonly less preserved than that learned more recently. In order to reduce the forgetting of particularly earlier learned old knowledge and improve the overall continual learning performance, we propose a simple yet effective fusion mechanism by including all the previously learned feature extractors into the intelligent model. In addition, a new feature extractor is included to the model when learning a new set of classes each time, and a feature extractor pruning is also applied to prevent the whole model size from growing rapidly. 
Experiments on multiple classification tasks show that the proposed approach can effectively reduce the forgetting of old knowledge, achieving state-of-the-art continual learning performance.
\end{abstract}


%
\IEEEpeerreviewmaketitle

\section{Introduction}

\IEEEPARstart{C}{ontinual} learning of new knowledge is one of the key abilities in human beings. With the ability, humans can accumulate their knowledge over time and become experts in certain domains. Such ability is also in high demand for intelligent classification systems considering that it is difficult, if not impossible, to collect training data for all classes at once in many application scenarios, such as in the applications of intelligent disease diagnosis and self-service store. Continual learning often presumes that very limited amount of or even no data is stored for old classes when learning new classes of knowledge each time due to various factors (e.g., privacy issue, no enough memory). This setting is also similar to the learning procedure of human beings, i.e., without reviewing too much old knowledge when learning new knowledge. However, when an intelligent model learns to acquire new classes of knowledge, it inevitably causes the change in model parameters, which then often leads to the rapidly decreased performance on previously learned old classes, knows as the \textit{catastrophic forgetting}~\cite{goodfellow2013empirical,kemker2018measuring}.

\begin{figure}[t]
\centering
\includegraphics[width=0.7\linewidth]{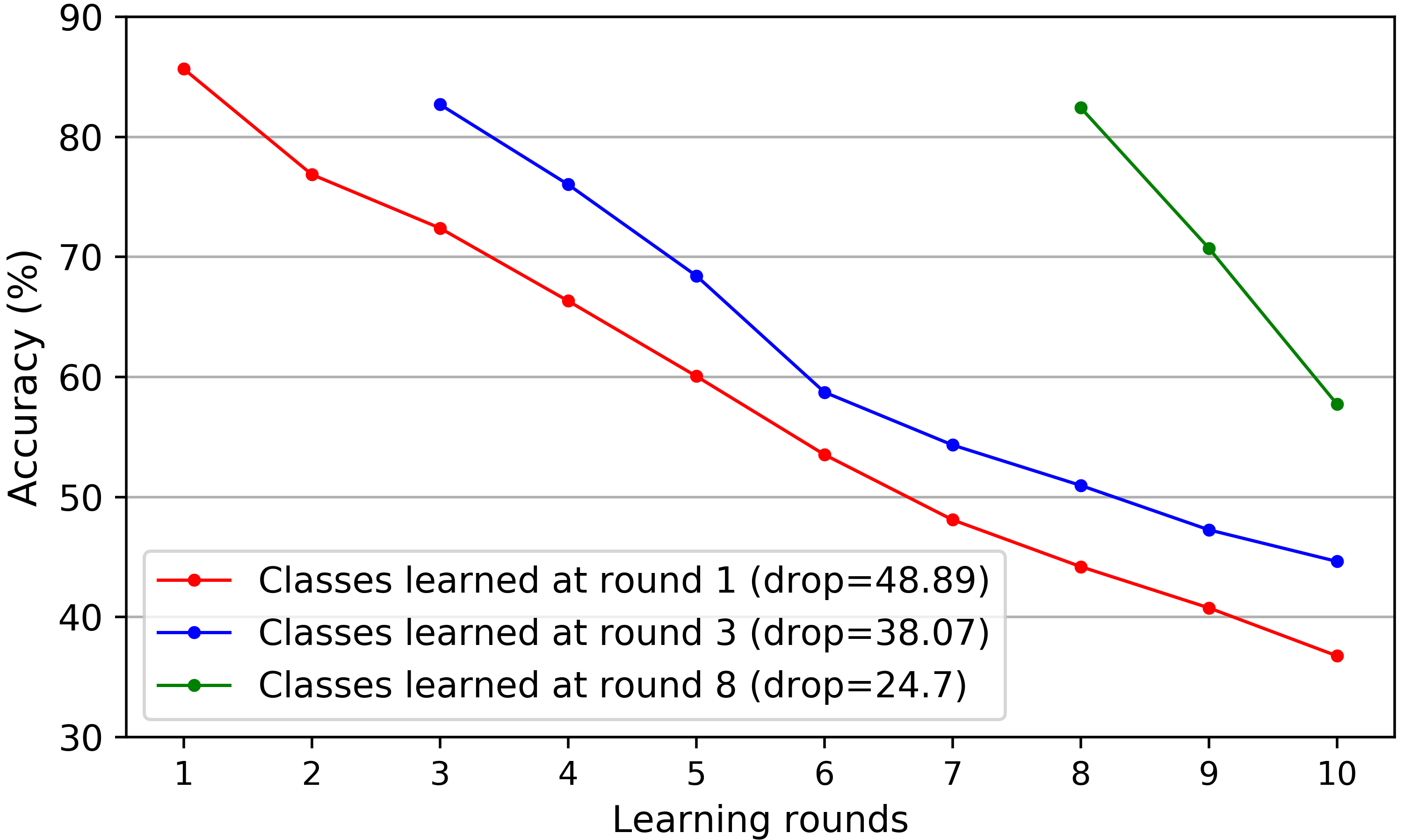}
\caption{Classification performance on three sets of old classes over rounds of continual learning. Knowledge of classes learned earlier (red) is forgotten more than those learned at later rounds (blue and green).
}
\label{fig:intro}
\end{figure}

To alleviate the catastrophic forgetting issue, state-of-the-art approaches (e.g., iCaRL~\cite{rebuffi2017icarl}, UCIR~\cite{hou2019learning}) often use stored limited amount of old classes' data and the model from the previous round of continual learning to help teach the new model, e.g., by distilling knowledge of old classes from the previous round of old model to the new model. While such approaches overall help reduce the forgetting of old knowledge, old knowledge learned at earlier rounds is often less preserved than that learned at recent rounds (Figure~\ref{fig:intro}). This is mainly because old knowledge is more or less forgotten at each round of learning and therefore old knowledge learned at earlier rounds would be forgotten more times than recently learned old knowledge. To alleviate the relatively more serious forgetting of old knowledge learned at earlier rounds, and therefore to further improve the overall performance of continual learning, it is probably necessary to learn not only from the most recent model but also from all earlier models when learning new knowledge each time.
 
In this study, we propose a simple yet effective continual learning framework which can learn old knowledge from all previous models during learning new classes. With the assumption that old knowledge learned at each round of continual learning is largely preserved in the feature extractor part of an convolutional neural network (CNN), where the CNN is particularly trained for the classes learned at that round, a knowledge fusion mechanism is proposed to embed the feature extractors of all previous rounds into the new classifier during learning new classes. All the previous feature extractors are fixed in the new classifier such that old knowledge learned at all previous rounds can be implicitly and partly preserved in the new classifier. Since the output space of different feature extractors are probably different, a learnable feature transformation follows each feature extractor and then all the transformed feature outputs are fused for the new classifier. Considering the model size would increase over rounds of continual learning due to the addition of one new feature extractor each time, a feature extractor pruning mechanism is applied to the newly trained feature extractor. Experiments on three classification datasets with different CNN backbones confirm that the proposed framework can better preserve old knowledge and achieve state-of-the-art performance.

\section{Related work}

There are two types of continual learning problems, task-incremental and class-incremental. Task-incremental learning presumes that one model is incrementally updated to solve more and more tasks, often with multiple tasks sharing a common feature extractor but having task-specific classification heads. The task identification is available during inference, i.e., users know which model head should be applied when predicting the class label of a new data. In contrast, class-incremental learning presumes that one model is incrementally updated to predict more and more classes, with all classes sharing a single model head. This study focuses on the class-incremental learning problem. Existing approaches to the two types of continual learning can be roughly divided into four groups, regularization-based, expansion-based, distillation-based, and regeneration-based.

Regularization-based approaches often find the model parameters or components (e.g., kernels in CNNs) crucial for old knowledge, and then try to keep them unchanged with the help of regularization loss terms when learning new classes~\cite{abati2020conditional,ahn2019uncertainty,fernando2017pathnet,jung2020continual,kim2018keep,kirkpatrick2017overcoming,mallya2018packnet}. The importance of each model parameter can be measured by the sensitivity of the loss function to changes in model parameters as in the elastic weight consolidation (EWC) method~\cite{kirkpatrick2017overcoming}, or by the sensitivity of the model output to small changes in model parameter as in the memory aware synapses (MAS) method~\cite{aljundi2018memory}. The importance of each kernel in a CNN model can be measured based on the magnitude of the kernel (e.g., L2 norm of the kernel matrix) as in the PackNet~\cite{mallya2018packnet}.  
While regularization-based approaches could help models keep old knowledge at first few rounds of continual learning where no much new knowledge need to be learned, it would become more difficult to continually learn new knowledge under the condition of keeping old knowledge because more and more kernels in CNNs become crucial for increasingly old knowledge. 

To make models more flexibly learn new knowledge, expansion-based approaches are developed by adding new kernels, layers, or even sub-networks when learning new knowledge~\cite{aljundi2017expert,hung2019compacting,karani2018lifelong,li2019learn,rajasegaran2019random,yoon2017lifelong}. 
For example, Aljundi et al.~\cite{aljundi2017expert} proposed expanding an additional network for a new task and training an expert model to make decisions about which network to use during inference. It turns a class-incremental learning problem to a task-incremental problem at the cost of additional parameters. As another example, Yoon et al.~\cite{yoon2017lifelong} proposed a dynamically expandable network (DEN) by selectively  retraining the network and expanding kernels at each layer if necessary. Most expansion-based and regularization-based approaches are initially proposed for task-incremental learning, although some of them (e.g., EWC) can be extended for class-incremental learning.

In comparison, distillation-based approaches can be directly applied to class-incremental learning by distilling knowledge from the old classifier (for old classes) to the new classifier (for both old and new classes) during learning new classes~\cite{castro2018end,hou2018lifelong,iscen2020memory,li2017learning,meng2020adinet,rebuffi2017icarl}, where the old knowledge is often implicitly represented by soft outputs of the old classifier with stored small number of old images and/or new classes of images as the inputs. A distillation loss is added to the original cross-entropy loss during learning the new classifier, where the distillation loss helps the new classifier have similar relevant output compared to the output of the old classifier for any input image. The well-known methods include the learning without forgetting (LwF)~\cite{li2017learning}, the incremental classifier and representation learning (iCaRL)~\cite{rebuffi2017icarl}, and the end-to-end incremental learning (End2End)~\cite{castro2018end}. More recently, the distillation has been extended to intermediate CNN layers, either by keeping feature map activations unchanged as in the learning without memorizing (LwM)~\cite{dhar2019learning}, or by keeping the spatial pooling unchanged respectively along the horizontal and the vertical directions as in PODNet~\cite{douillard2020podnet}, or by keeping the normalized global pooling unchanged at last convolutional layer as in the unified classifier incrementally via rebalancing (UCIR)~\cite{hou2019learning}. These distillation-based methods achieve the state-of-the-art performance for the class-incremental learning problem.

In addition, regeneration-based approaches have also been proposed particularly when none of old-class data is available during learning new classes. The basic idea is to train an auto-encoder~\cite{hayes2019remind,rao2019continual,riemer2019scalable} or generative adversarial network (GAN)~\cite{ostapenko2019learning,rios2018closed,shin2017continual,xiang2019incremental} to synthesize old data for each old class, such that plenty of synthetic but realistic data for each old class together with new classes of data can be used to train the new classifier. However, such methods heavily depends on the quality of synthetic data, and data quality could become worse especially when using a single regeneration model for more and more old classes.

\section{Problem description}

The problem of interest is to incrementally learn more and more classes over time. Formally, for the $t$-th round of continual learning, suppose $c_t$ new classes of training dataset $D_t$ is available to update the previously trained classifier $\mathcal{M}_{t-1}$, and the new updated classifier $\mathcal{M}_t$ is able to recognize $c_1+c_2+\ldots+c_t$ classes of data. The classifier $\mathcal{M}_1$ at the first round (i.e., when $t=1$) is often trained from the scratch based on the first dataset $D_1$. While it would be ideal to update the classifier based on only the new classes of dataset $D_t$ and without using any previous old datasets $\{D_1, D_2,\ldots, D_{t-1}\}$ at the $t$-th round of learning, recent studies~\cite{hou2019learning,rebuffi2017icarl} have shown that a small subset of data for each old class is necessary to effectively help keep the new classifier from catastrophic forgetting of old knowledge. Similarly, in this study, it assumes that the new dataset and very small subsets of old data, together $\mathcal{D}_t = \{D'_1, D'_2,\ldots, D'_{t-1}, D_t\}$, are available to update the classifier at the $t$-th round of continual learning. Overall, the objective of this study is to develop a learning system $\mathcal{S}$ which can effectively update the previously learned classifier $\mathcal{M}_{t-1}$ based on the available dataset $\mathcal{D}_t$, i.e.,
\begin{equation}
\mathcal{M}_t =\mathcal{S} (\mathcal{M}_{t-1}, \mathcal{D}_t) \,,
\end{equation}
such that the updated classifier $\mathcal{M}_t$ can accurately recognize both $c_1+c_2+\ldots+c_{t-1}$ old  and  $c_t$ new classes.

\section{Our approach}

\begin{figure*}[!tbh]
\centering
\includegraphics[width=0.9\linewidth]{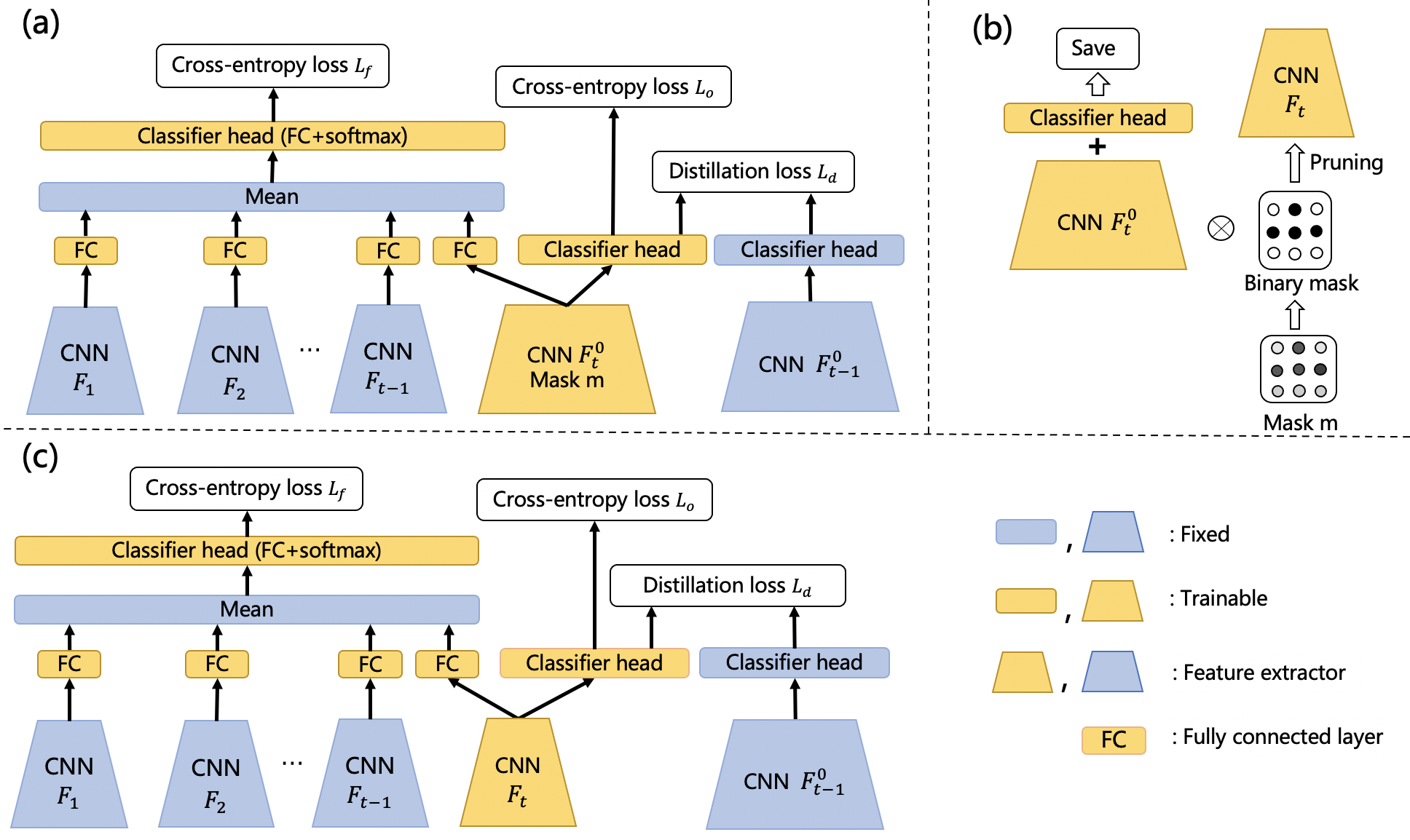}
\caption{The proposed continual learning framework. The classifier consists of multiple feature extractors and one single classifier head. (a) The new feature extractor $F_{t}^0$ at current round of continual learning is learned partly with the help of knowledge distillation from the last-round feature extractor $F_{t-1}^0$. All previous and current feature extractors are fused as part of the updated classifier. A mask is also learned to measure the importance of each kernel in the new feature extractor $F_{t}^0$. (b) The new feature extractor is pruned based on the learned kernel mask. (c) The whole system is fine-tuned after replacing $F_{t}^0$ by its thinner version $F_{t}$.
}
\label{fig:whole_procedure}
\end{figure*}

The key challenge of continual learning is to alleviate the issue of catastrophic forgetting, i.e., trying to preserve the knowledge of old classes when learning new classes. Although a small subset of data has been kept for each old class and knowledge distillation from the previous classifier $\mathcal{M}_{t-1}$ to the new classifier $\mathcal{M}_t$ has helped the new classifier keep old knowledge to some extent, the dominant new classes of data during classifier updating would make the new classifier oriented to these new classes during inference~\cite{belouadah2019il2m,wu2019large,zhao2020maintaining}. If the classifier $\mathcal{M}_{t-1}$ is a CNN model, classifier updating often modifies not only the classifier's head structure (by adding more output neurons for new classes) but also the parameters of all CNN layers. The majority of a CNN classifier is composed of multiple convolutional layers which can be considered as a feature extractor for the final classifier's head. Since knowledge of each old class is largely and implicitly preserved in the feature extractor of the corresponding old classifier (by which the old class is first learned), the parameter changes in the feature extractor at each round of continual learning would inevitably lead to forgetting of old knowledge. With this consideration, we propose combining the feature extractors learned from all previous classifiers with the current feature extractor for the updated new classifier $\mathcal{M}_t$. The feature extractor learned at each previous round is fixed and becomes part of the current new classifier (Figure~\ref{fig:whole_procedure}a). Since different feature extractors are learned at different time and incrementally for more and more classes, they would not share the same output feature space. Therefore, directly combining the output of all feature extractors is not meaningful and may not effectively help preserve old knowledge. Here we propose learning to transform the output of each feature extractor into one common feature space and then combining the transformed features for final classification. Each to-be-learned transformation can be realized by a fully connected layer following each feature extractor (Figure~\ref{fig:whole_procedure}, FC in yellow). The new feature extractor $F_t$ specifically for the current classifier can be jointly learned, e.g., with the knowledge distillation mechanism. In addition, considering the classifier would become larger with more rounds of continual learning due to the addition of all old feature extractors, we also propose pruning each newly learned feature extractor and only the pruned feature extractors are embedded into subsequent new classifiers. The detailed classifier updating process is described below. 

\subsection{Knowledge distillation}

A new (pruned) feature extractor $F_t$ as part of the updated new classifier $\mathcal{M}_t$ would be learned at the current $t$-th round of continual learning (Figure~\ref{fig:whole_procedure}). Similar to the continual learning approaches iCaRL~\cite{rebuffi2017icarl} and End2End~\cite{castro2018end}, the knowledge distillation mechanism is adopted to help train the feature extractor. Suppose the feature extractor $F_t$ is a pruned version of an original feature extractor $F^0_t$ which contains more kernels than $F_t$ at each convolutional layer (see Section~\ref{sec:pruning} for pruning detail), and $F^0_t$ together with its specific classifier head form an individual CNN classifier $M^0_t$ (yellow part on the right side in Figure~\ref{fig:whole_procedure}a). $M^0_t$ is a temporary classifier at the $t$-th round  for both old and new classes and will be jointly trained with the other parts of the classifier $\mathcal{M}_t$, with the help of knowledge distillation from the corresponding individual classifier $M^0_{t-1}$ (blue part on the right side in Figure~\ref{fig:whole_procedure}a) learned at the previous round of continual learning.
Specifically, for every training data $(\ve{x}_i, \ve{y}_i) \in  \mathcal{D}_t$ where $\ve{x}_i$ is the $i$-th training image in $\mathcal{D}_t$ and $\ve{y}_i$ is one-hot vector, denote by $\textbf{z}_i = [z_{i1}, z_{i2}, \cdots, z_{iu}]^{\mathsf{T}}$ the logit output (before performing the softmax operation) of the previous classifier $M^0_{t-1}$, where $u$ is the total number of old classes ($u=c_1+c_2+\ldots+c_{t-1}$). Similarly denote by $\hat{\textbf{z}}_i = [\hat{z}_{i1}, \hat{z}_{i2}, \cdots, \hat{z}_{i,u+c_t}]^{\mathsf{T}}$ the logit output of $M^0_t$, where $c_t$ is the number of new classes. 
Then, the distillation of old knowledge from $M^0_{t-1}$ to $M^0_t$ can be realized by the minimization of the distillation loss $\mathcal{L}_{d}$,
\begin{equation}
\mathcal{L}_{d}(\vv{\theta}) = -\frac{1}{N} \sum_{i=1}^N \sum_{j=1}^u p_{ij} \log \hat{p}_{ij}  \,,
\end{equation}
where $\vv{\theta}$ represents the to-be-learned parameters of  $M^0_t$, N is the total number of training data in $\mathcal{D}_t$, and $p_{ij}$ and $\hat{p}_{ij}$ are from the modified softmax operation,
\begin{equation}
p_{ij}= \frac{\exp{(z_{ij}/T)}}{\sum_{k=1}^u \exp{(z_{ik}/T)}} \,, \quad
\hat{p}_{ij}= \frac{\exp{(\hat{z}_{ij}/T)}}{\sum_{k=1}^u \exp{(\hat{z}_{ik}/T)}} \,,
\end{equation}
and $T\ge 1$ is the distillation parameter. Besides, the cross-entropy loss $\mathcal{L}_{o}$ is also applied to help $M^0_t$ discriminate both old and new classes,
\begin{equation} \label{eq:ce}
\mathcal{L}_{o}(\vv{\theta}) = - \frac{1}{N} \sum_{i=1}^N \sum_{k=1}^{u+c_t} y_{ik} \log (p_{ik}) \,,
\end{equation}
where $y_{ik}$ is the $k$-th element of the one-hot class label vector $\ve{y}_i$ and $p_{ik}$ is the $k$-th element of the softmax output from the individual classifier $M^0_t$. In combination, $M^0_t$ can be trained partly by minimizing the loss $\mathcal{L}_{s}$,
\begin{equation} \label{eq:integrated_loss_single}
\mathcal{L}_{s}(\vv{\theta}) = \mathcal{L}_{o}(\vv{\theta}) + \lambda_1 \mathcal{L}_{d}(\vv{\theta}) \,,
\end{equation}
where $\lambda_1$ is a trade-off coefficient to balance the importance of the two loss terms.

\subsection{Knowledge fusion}
As mentioned above, the new feature extractor $F^0_t$ (which is finally replaced by its pruned version $F_t$) in the individual CNN classifier $M^0_t$ is combined with all previously learned old feature extractors $\{F_1, F_2, \ldots, F_{t-1}\}$ to form part of the feature extractor of the new classifier $\mathcal{M}_t$ (Figure~\ref{fig:whole_procedure}a). Every old feature extractor is mainly responsible for knowledge preservation of a specific subset of old classes, and therefore the combination of both old feature extractors and the new feature extractor would fuse knowledge of all old and new classes. Note that all old feature extractors are fixed during training the new classifier $\mathcal{M}_t$ and only the subsequent fully connected layers and the new feature extractor are learnable in the new classifier. Similar to Equation~\ref{eq:ce}, another cross-entropy loss $L_f$ is used to measure the difference between the outputs of the updated classifier $\mathcal{M}_t$ and the ground-truth one-hot vectors for all training data. Since the new feature extractor is shared by both the new classifier  $\mathcal{M}_t$ and the individual CNN classifier $M^0_t$, the two classifiers are jointly trained by combining all the above loss functions, i.e., by minimizing the loss $\mathcal{L}$,
\begin{equation}
\mathcal{L} = \mathcal{L}_{f} + \lambda_2 \mathcal{L}_s \,,\label{eqn:loss}
\end{equation}
where $\lambda_2$ is a trade-off coefficient.

\subsection{Feature extractor pruning}
\label{sec:pruning}

In order to prevent the rapid growing of the model size, the initial learned feature extractor $F^0_t$ is pruned to obtain the thinner version $F_t$ (Figure~\ref{fig:whole_procedure}b) and then the whole model is fine-tuned (Figure~\ref{fig:whole_procedure}c).
Inspired by PackNet~\cite{mallya2018packnet} and PathNet~\cite{fernando2017pathnet} which use network pruning for task-incremental learning based on a single CNN model, a real-valued mask for each convolutional layer is jointly learned during the training of the updated classifier $\mathcal{M}_t$ and the individual CNN classifier $M^0_t$ (Figure~\ref{fig:whole_procedure}a). Each component in the mask represents the importance of one unique kernel, and the kernel with relatively less importance would be pruned from each convolutional layer.
Formally, denote by $\ve{w}_{l,h}$ and $m_{l,h}$ the $h$-th kernel and the associated importance element in the mask $\ve{m}_l$ at the $l$-th layer, respectively. To assure each $m_{l,h}$ is within the range $[0,1]$, we propose that $\ve{m}_l$ is from a learnable vector $\ve{e}_l$ of the same dimensionality, with the deterministic relationship $\ve{m}_l = \text{softmax}(\ve{e}_l)$.
The mask $\ve{m}_l$ (actually $\ve{e}_l$) is embedded into the convolution process with 
\begin{equation}
\ve{f}_{l,h} = (m_{l,h} \ve{w}_{l,h}) \circledast \ve{f}_{l-1}  \,,
\end{equation}
where $\circledast$ represents the convolution operator, $\ve{f}_{l-1}$ is the input to the $l$-th layer (also the output of the previous layer), and $\ve{f}_{l,h}$ is the result of the convolution between the kernel $\ve{w}_{l,h}$ and the input, weighted by the importance $m_{l,h}$ of the kernel. Suppose there are totally $n_l$ kernels at the $l$-th layer, and each $m_{l,h}$ is initialized with $1/n_l$ (by initializing $\ve{e}_l = \ve{0}$).
After jointly training of all the masks $\{\ve{m}_l\}$, the new classifier  $\mathcal{M}_t$ and the individual CNN classifier $M^0_t$, a binary mask $\ve{\tilde{m}}_l$  at each layer can be obtained simply with the thresholding function 
\begin{equation}
\ve{\tilde{m}}_l = \mathbb{I}(\ve{m}_l \geq \frac{1}{n_l}) \,,
\end{equation}
considering that any value less than $1/n_l$ indicates that the corresponding kernel is less important. Note that adaptive threshold could be adopted for more flexible pruning. With the binary masks, associated less important kernels are pruned from the initially learned feature extractor $F^0_t$ at each layer (Figure~\ref{fig:whole_procedure}b). Following the general routine for network pruning, the pruned thinner feature extractor $F_t$, together with the other parts of the updated classifier  $\mathcal{M}_t$, is fine-tuned, again with the help of knowledge distillation from the individual CNN classifier $M^0_{t-1}$ (Figure~\ref{fig:whole_procedure}c). 

Note that one concurrent work, called dynamically expandable representation (DER)~\cite{yan2021dynamically}, has recently been released with a very similar idea. DER is also built on the fixed old feature extractors and the pruning of each new feature extractor. However, DER does not use knowledge distillation and the learnable fully connected layer following each old feature extractor, and the pruning procedure in DER is more complicated than ours. As shown in the experiments, the learnable fully connected layer in our framework can further improve the continual learning framework.

\section{Experiments}

\subsection{Experimental settings}

Our approach was evaluated on three datasets, CIFAR100~\cite{krizhevsky2009learning}, ImageNet~\cite{deng2009imagenet}, and a subset of ImageNet (Table~\ref{tab:dataset}). The subset of ImageNet (mini-ImageNet) was randomly selected from the original one, containing 100 classes totally.
In the training phase, each CIFAR100 image was randomly flipped horizontally before being input to the classifier, and each ImageNet image was randomly cropped and then resized to 224 $\times$ 224 pixels.
On each dataset, an CNN classifier was first trained for certain number (e.g., 10, 20) of classes, and then a set of new classes' data were provided to update the classifier at each round of continual learning. 
Adam optimizer (batch size 128) was used with an initial learning rate 0.001. Each model was trained for up to 100 epochs, with the training convergence consistently observed. 
ResNet18 was used as the default CNN backbone, and $\lambda_1=1.0$, $\lambda_2=0.1$. Following  iCaRL~\cite{rebuffi2017icarl}, the herding strategy was adopted to select a small subset of images for each new class with a total memory size 2000, and the nearest-mean-of-exemplars (NME) method was adopted during testing.

\begin{table}[!tbh]
    \centering
    \caption{Statistics of two datasets used in experiments. [75, 2400] represents the range of image height and width. }
    \begin{tabular}{ccx{1.45cm}x{1.45cm}c}
        \toprule
         Dataset  & Classes & \# Train \ \ \ (per class) & \# Test \ \ \ (per class)   &  Size \\ 
         \midrule
         CIFAR100      & 100 & 500    & 100  & 32$\times$32  \\
         mini-ImageNet & 100 & 1,000  & 100  & [75, 2400]  \\
         ImageNet      & 1,000 & $\sim$1,200 & 50  & [75, 2400]  \\
         \bottomrule
    \end{tabular}
    \label{tab:dataset}
\end{table}

After training at each round, the average accuracy over all learned classes so far was calculated. 
Such a training and evaluation process was repeated in next-round continual learning. For each experiment, the average and standard deviation of accuracy over five runs were reported, each run with a different and fixed order of classes to be learned. All baseline methods were evaluated on the same orders of continual learning over five runs and with the same herding strategy for sample selection.

\begin{figure*}[!bht]
    \centering
    \includegraphics[width=0.30\linewidth]{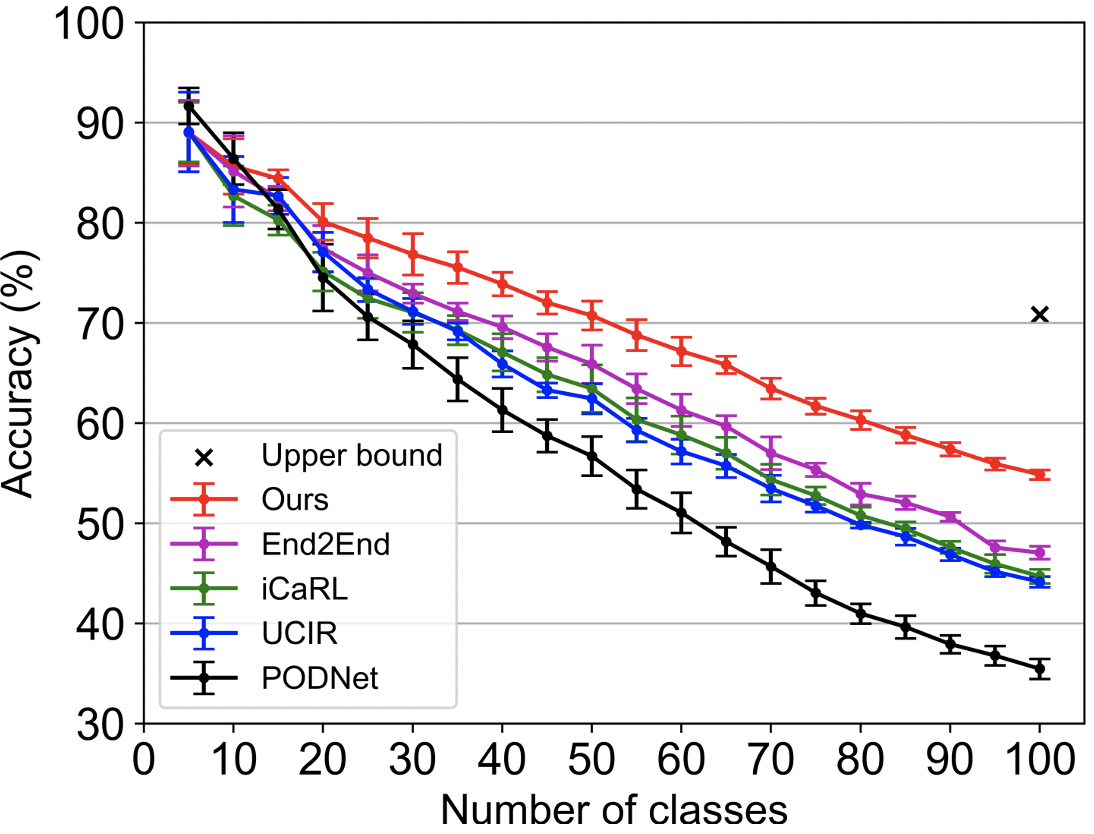}
    \includegraphics[width=0.30\linewidth]{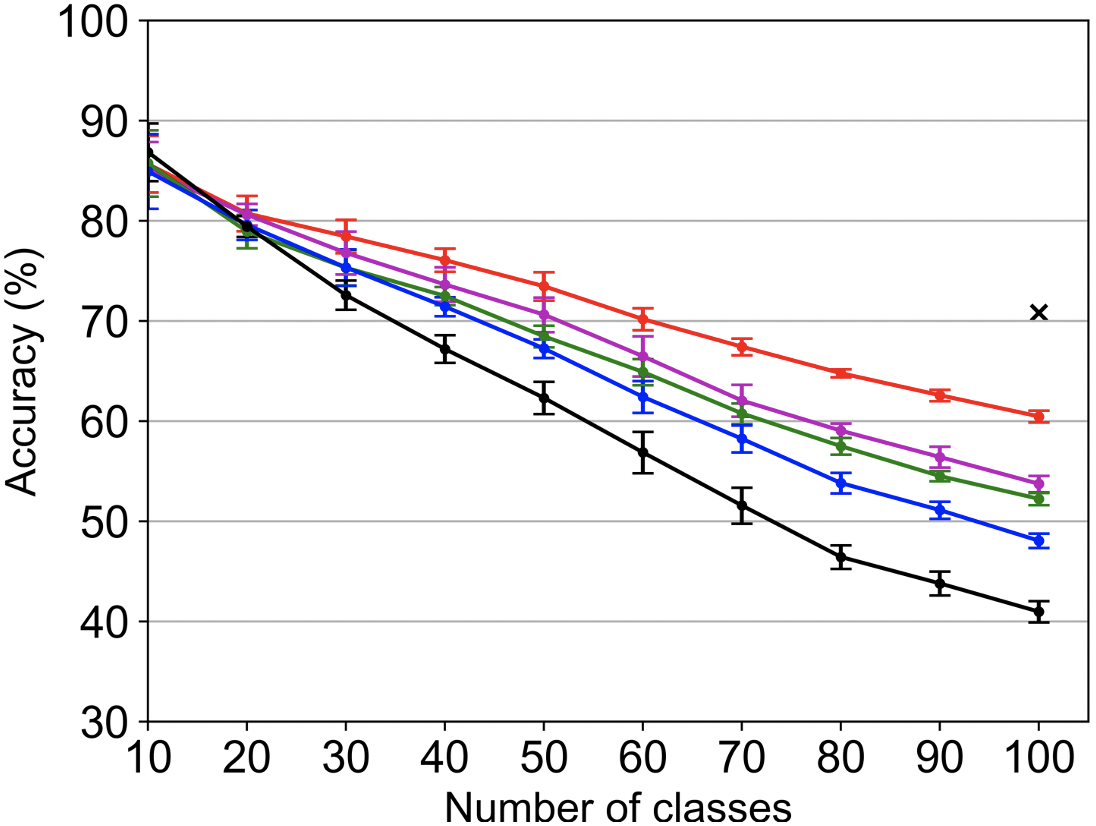}
    \includegraphics[width=0.30\linewidth]{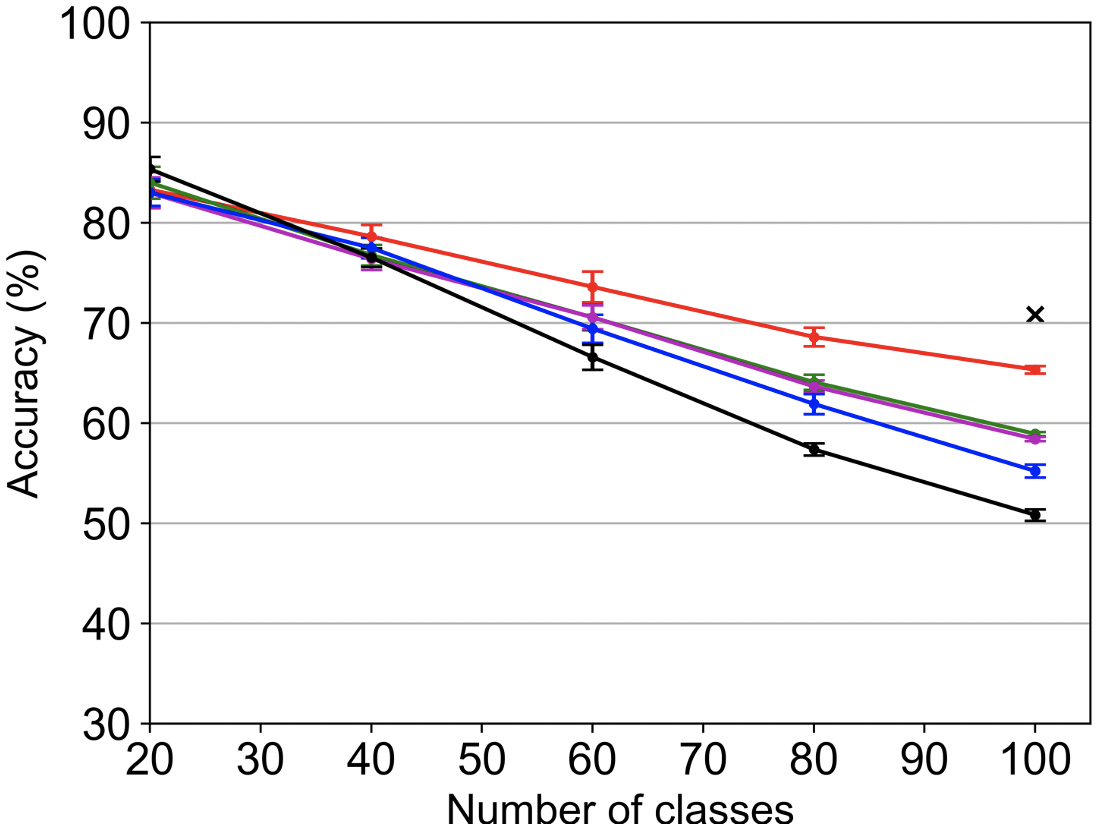}
    \\
    \includegraphics[width=0.30\linewidth]{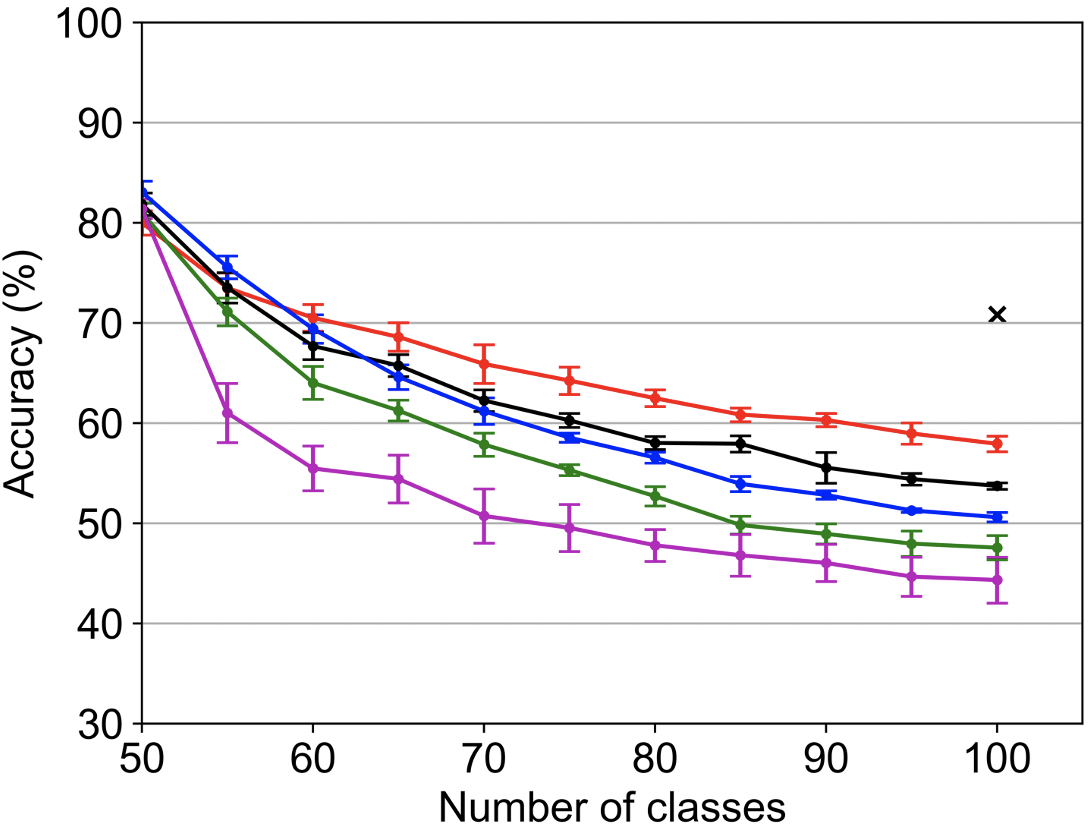}
    \includegraphics[width=0.30\linewidth]{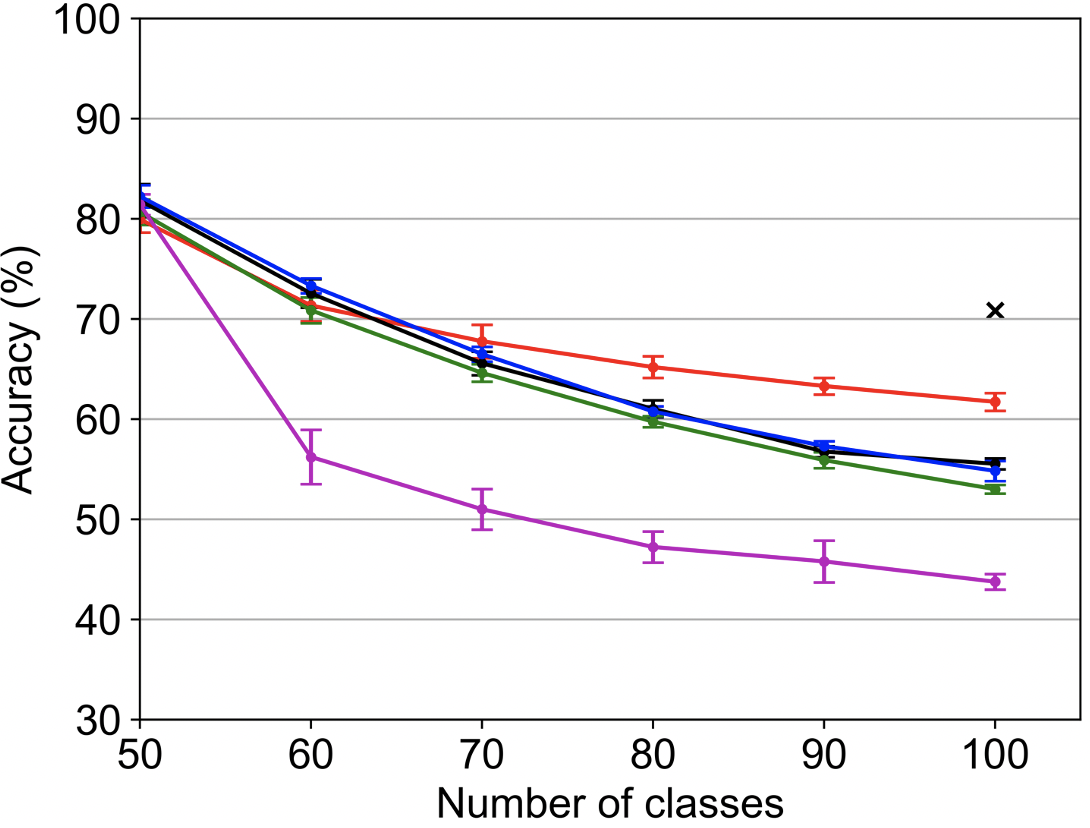}
    \includegraphics[width=0.3\linewidth]{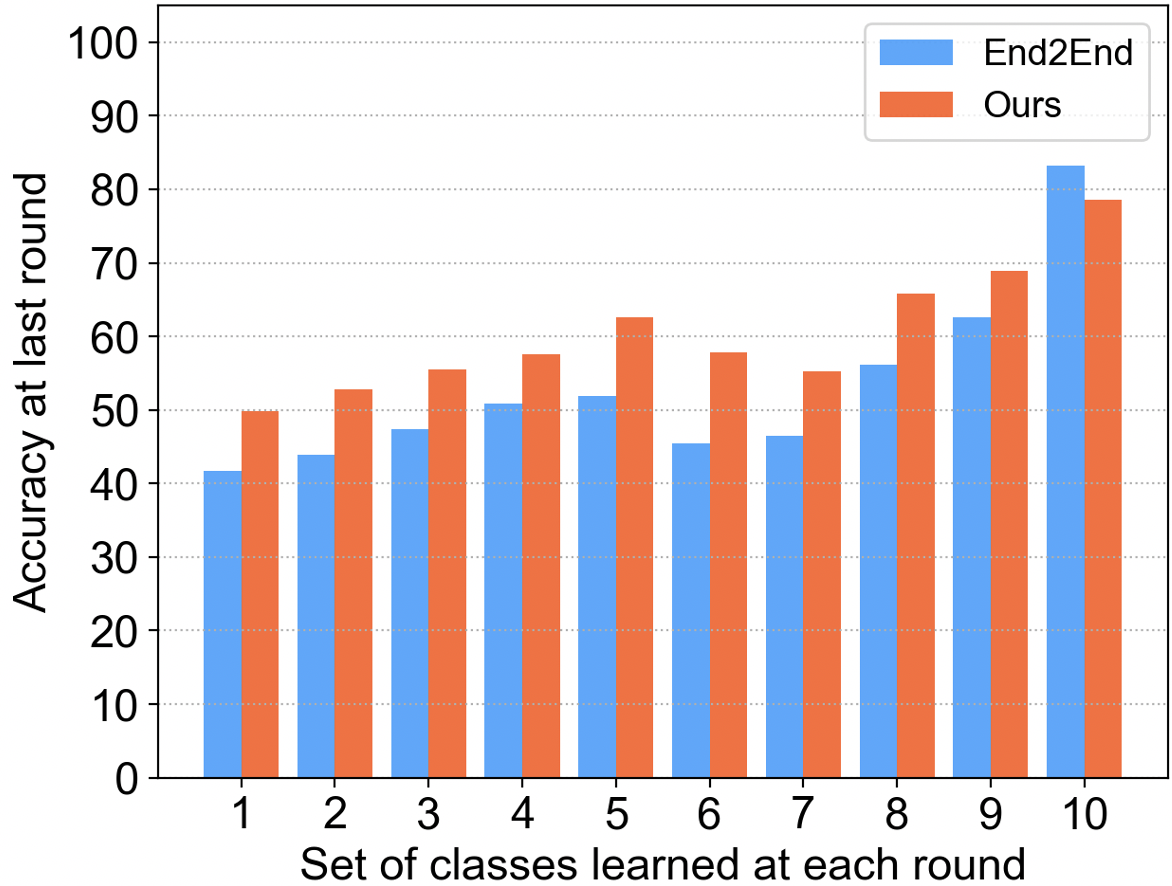}
\caption{Performance comparison between our approach and the state-of-the-art methods on the CIFAR100 dataset. The upper bound result (black cross) was obtained by training a single CNN classifier with all the training data of the 100 classes.}
\label{fig:cifar100}
\end{figure*}

\subsection{Effectiveness evaluation}

\begin{figure*}[!tbh]
 \centering
 \captionsetup{justification=centering}
 \includegraphics[width=0.30\linewidth]{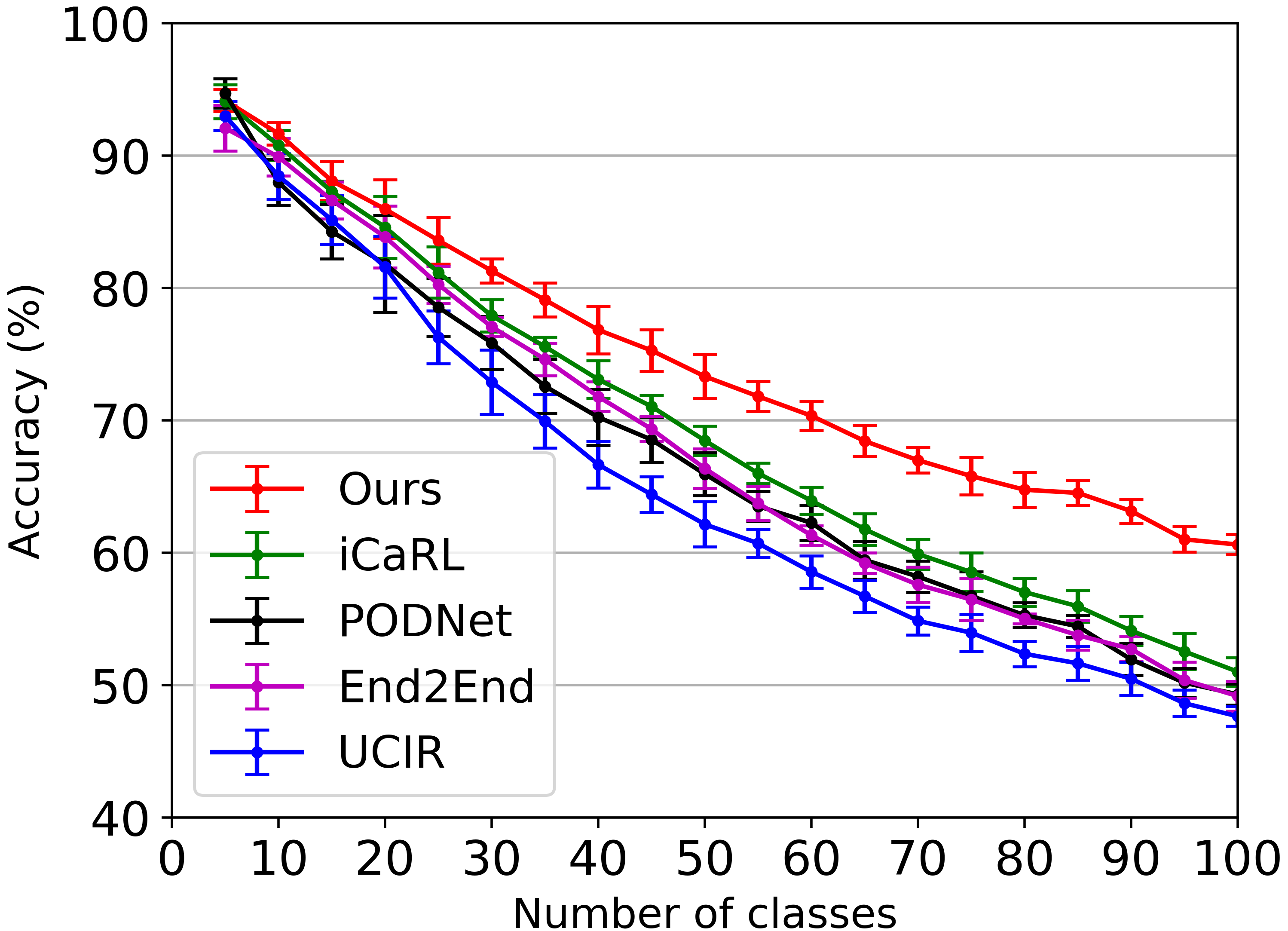}
 \includegraphics[width=0.30\linewidth]{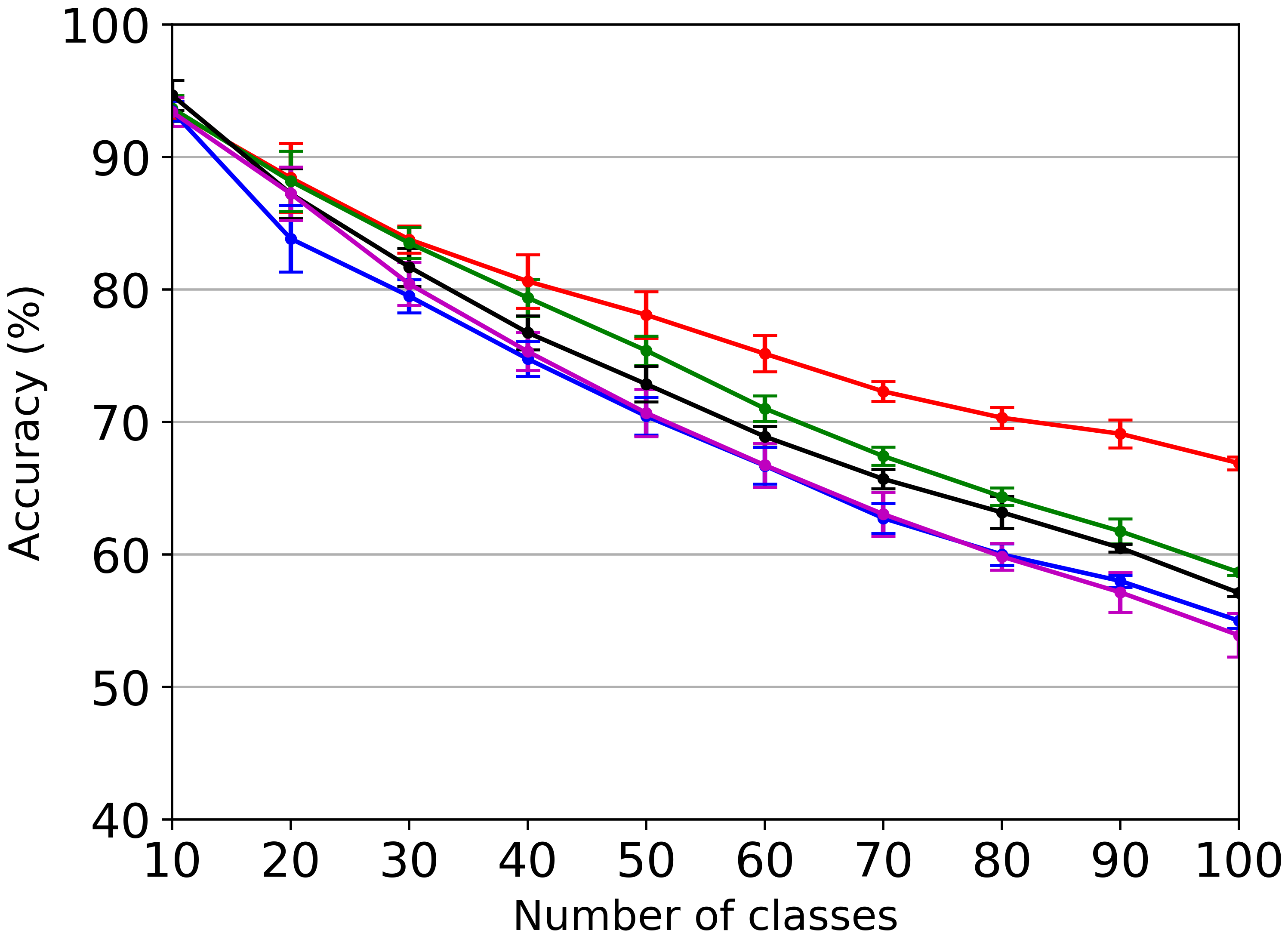}
 \includegraphics[width=0.30\linewidth]{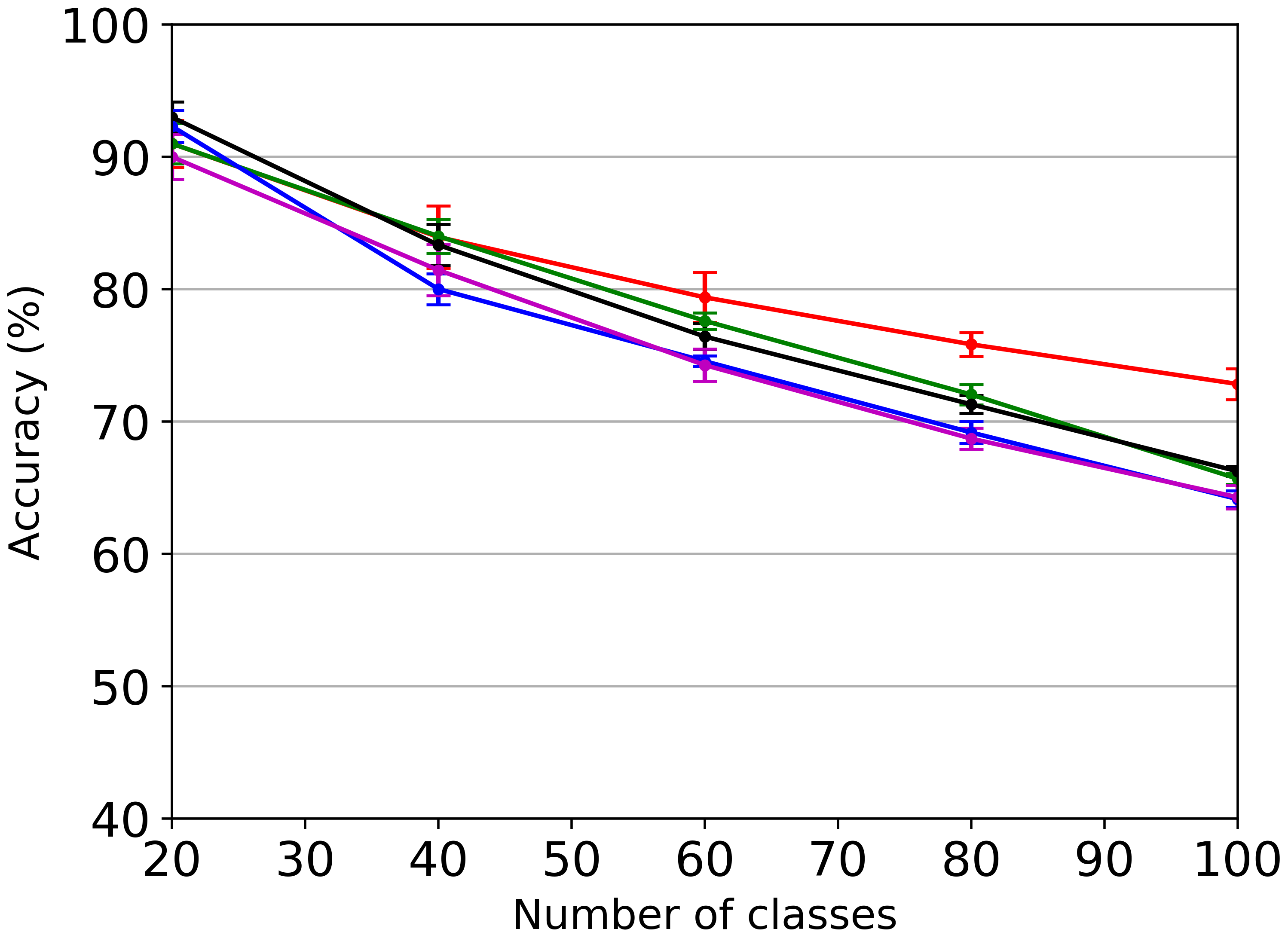}
\caption{Performance comparison between our approach and the state-of-the-art methods on the mini-ImageNet dataset.}
\label{fig:mini_imagenet}
\end{figure*}

\begin{table*}[!tbh]
\centering
\caption{Continual learning performance with different CNN backbones. }
\begin{tabular}{cx{0.95cm}x{0.95cm}x{0.95cm}x{0.95cm}x{0.95cm}x{0.95cm}x{0.95cm}x{0.95cm}x{0.95cm}x{0.95cm}}
\toprule
Backbone & \multicolumn{2}{c}{iCaRL} & \multicolumn{2}{c}{End2End} & \multicolumn{2}{c}{UCIR} & \multicolumn{2}{c}{PODNet} & \multicolumn{2}{c}{Ours} \\
\cmidrule(l){2-3}
\cmidrule(l){4-5}
\cmidrule(l){6-7}
\cmidrule(l){8-9}
\cmidrule(l){10-11}
         & ACC & Mean & ACC & Mean & ACC & Mean & ACC & Mean & ACC & Mean \\
\midrule
WRN16  & 50.5 $\pm$ \small{0.9} & 64.4 $\pm$ \small{1.0} & 53.0 $\pm$ \small{0.9} & 65.3 $\pm$ \small{1.2} & 51.1 $\pm$ \small{0.9} & 64.0 $\pm$ \small{1.1} & 45.6 $\pm$ \small{1.4} & 60.5 $\pm$ \small{1.5} & \textbf{62.3} $\pm$ \small{0.4} & \textbf{71.0} $\pm$ \small{0.8}  \\
\midrule
SENet18  & 50.7 $\pm$ \small{1.0} & 64.2 $\pm$ \small{1.3} & 45.8 $\pm$ \small{0.7} & 61.3 $\pm$ \small{0.9} & 48.8 $\pm$ \small{0.9} & 63.2 $\pm$ \small{1.0} & 44.4 $\pm$ \small{0.8} & 59.4 $\pm$ \small{0.8} & \textbf{58.2} $\pm$ \small{0.4} & \textbf{68.2} $\pm$ \small{0.7}  \\
\midrule
ResNet34  & 51.2 $\pm$ \small{0.9} & 64.6 $\pm$ \small{1.0} & 52.4 $\pm$ \small{0.5} & 65.6 $\pm$ \small{0.9} & 49.9 $\pm$ \small{0.5} & 64.4 $\pm$ \small{0.7} & 45.0 $\pm$ \small{1.0} & 59.7 $\pm$ \small{0.6} & \textbf{61.7} $\pm$ \small{0.6} & \textbf{70.7} $\pm$ \small{0.7}  \\ 
\bottomrule
\end{tabular}
\label{table:diff_arch}
\end{table*}

The effectiveness of our approach was evaluated firstly on the CIFAR100 dataset, in comparison with the state-of-the-art methods iCaRL~\cite{rebuffi2017icarl}, End2End~\cite{castro2018end}, UCIR~\cite{hou2019learning}, and PODNet~\cite{douillard2020podnet}.
PODNet was run with the source code released by its authors, and the other baselines were re-implemented and similar amount of effort was put into tuning each baseline method. 
As shown in Figure~\ref{fig:cifar100}, when continually learning 5 classes (First row, left), 10 classes (First row, middle), and 20 classes (First row, right) at each round respectively, our approach clearly outperforms all these state-of-the-art methods, with at least absolute $5\%$ better than all others in accuracy at the final round of learning. Figure~\ref{fig:cifar100} (Second row, right) also shows that, compared to the best baseline method End2End (blue) when learning 10 new classes at each round, the final-round classifier by our approach (orange) performs better for each set of old classes learned at each round, confirming that our approach can better preserve the knowledge of old classes learned at earlier rounds. 

Considering that some methods (e.g., PODNet, UCIR) were originally reported based on an initial classifier which was trained on 50 classes, similar experiments were performed here for more comprehensive evaluation. Figure~\ref{fig:cifar100} (Second row, left and middle) again shows that our approach achieves the best performance when continually learning 5 or 10 classes each time, better than the best baseline PODNet by absolute $4\%$ in accuracy at the last round. 
One interesting observation is that, while PODNet performs equivalently well or better than the other strong baselines under the condition of starting from a 50-class classifier (Second row, left and middle), it is worse than the others when starting from a 5-class, 10-class, or 20-class classifier (First row).
Such result indicates that performance of these existing state-of-the-art methods may be sensitive to the initially learned amount of knowledge. In particular, PODNet may work well only when enough amount of knowledge is initially learned before starting to learn new classes.
In comparison, our approach is more robust to the amount of knowledge learned at the first round and consistently better than the others under various learning conditions.

\begin{figure}[!tbh]
 \centering
 \includegraphics[width=0.35\textwidth]{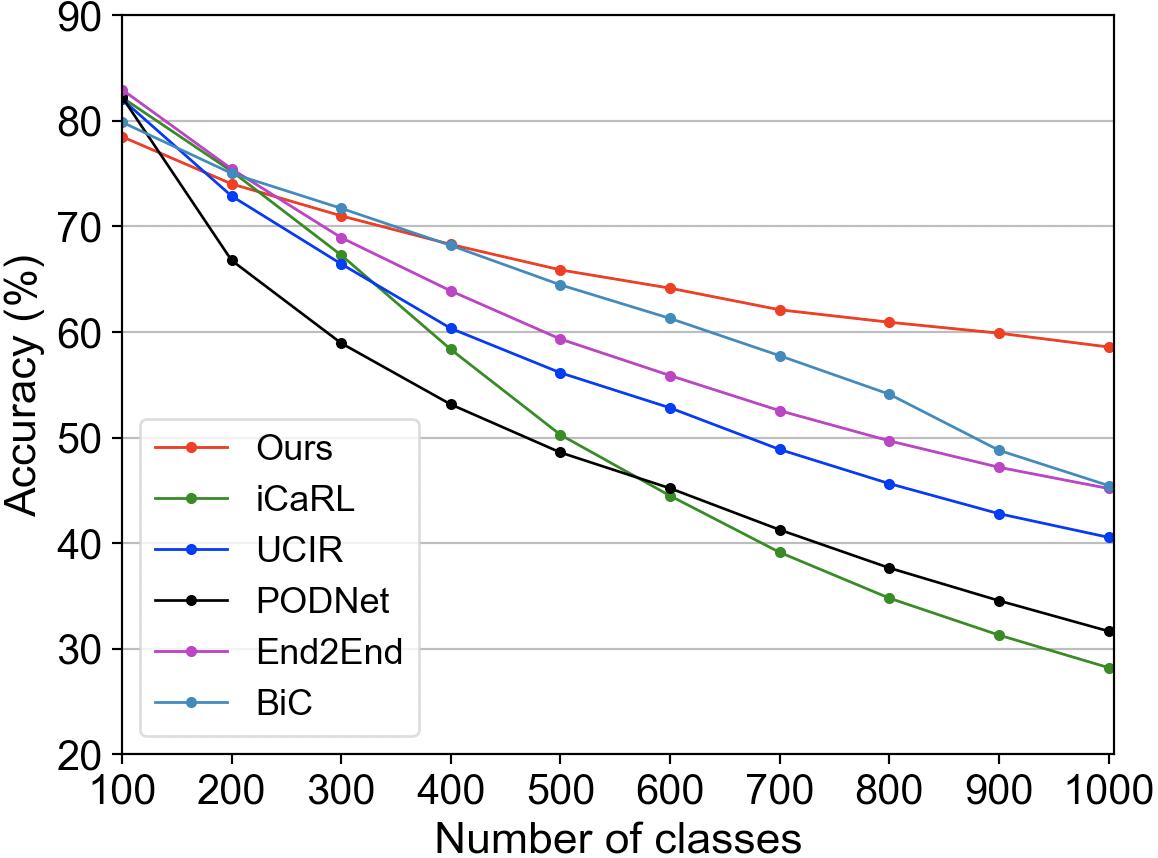}
\caption{Performance comparison between our approach and the state-of-the-art methods on the ImageNet dataset.}
\label{fig:imagenet}
\end{figure}

Similar results were obtained from experiments on the mini-ImageNet dataset (Figure~\ref{fig:mini_imagenet}) and the ImageNet dataset (Figure~\ref{fig:imagenet}). One additional strong baseline BiC~\cite{wu2019large} which was designed particularly for large-scale incremental learning was also included in the comparison on ImageNet. Again, the state-of-the-art methods UCIR and PODNet perform relatively poor when the number of initial classes is small, and our approach is consistently better than all strong baselines. The gap (at least $7\%$ on mini-ImageNet and $13\%$ on ImageNet) at the final round between the baselines and our approach is even larger than that on CIFAR100.

\subsection{Evaluation on different backbones}

To demonstrate the generalizability of our approach, multiple CNN backbones were respectively used for feature extractors in the framework, including the 16-layer WideResNet (WRN16)~\cite{zagoruyko2016wide}, the 18-layer SENet~\cite{hu2018senet}, and the 34-layer ResNet (ResNet34). Experiments were performed on the CIFAR100 data with continual learning of 10 classes at each round. The classification accuracy (ACC) at the final round, the mean of classification accuracy (Mean) over all but the initial rounds, and their standard deviations over five runs were reported.  
As shown in Table~\ref{table:diff_arch}, among the baselines, PODNet still performs worst probably due to the limited amount of initial knowledge, and the End2End is unusually sensitive to the network backbones. While the performance varies a bit over different backbones for each approach, our approach consistently outperforms the baselines with all CNN backbones, suggesting the good generalizability of the proposed framework.

\subsection{Ablation study}
\label{sec:ablation}

To evaluate the effect of key components in the proposed framework, a series of tests were performed by progressively removing part(s) of the framework, including the removal of the feature extractor pruning (Mask in Table~\ref{ablation:stack}), the trainable fully connect layers for each feature extractor (FC in Table~\ref{ablation:stack}), and the fixed feature extractors learned at previous rounds (Fusion in Table~\ref{ablation:stack}).
As shown in Table~\ref{ablation:stack}, the inclusion of the fixed feature extractors only (Second row)  already improves the classification performance by 5\% in accuracy at the final round, compared to the single CNN model (First row). The inclusion of the trainable fully-connected layer for each feature extractor (Third row) further improves the performance by about 3\%, confirming that transformations are necessary to map features from different feature extractors to the same feature space for better knowledge fusion. Interestingly, the feature extractor pruning not only reduces the model size but also has a positive impact on knowledge preservation (Last row vs. Third row). One possible reason is that the pruned feature extractors, or the smaller model, contain fewer model parameters and therefore is less likely over-fitting with better generalization. It is worth noting that the proposed approach outperforms the strong baselines (Figure~\ref{fig:cifar100}) even without pruning feature extractors.
These results demonstrate that each part of the proposed framework is necessary and effective.

\begin{table}[t]
\centering
\caption{The effect of each framework component, with continual learning of 10 classes at each round on CIFAR100.}
\begin{tabular}{x{1cm}x{0.6cm}x{0.8cm}x{1.5cm}x{1.7cm}}
\toprule
Fusion & FC & Mask & ACC & Mean  \\ 
\midrule
    ---     &     ---    &     ---    &  50.7 $\pm$ 0.9  &  64.5 $\pm$ 1.0  \\ 
\checkmark  &     ---    &     ---    &  56.0 $\pm$ 0.7  &  67.9 $\pm$ 0.7  \\
\checkmark  & \checkmark &     ---    &  58.5 $\pm$ 0.4  &  69.3 $\pm$ 0.8  \\
\checkmark  & \checkmark & \checkmark &  \textbf{60.5} $\pm$ 0.6  &  \textbf{70.5} $\pm$ 0.8  \\ 
\bottomrule
\end{tabular}
\label{ablation:stack}
\end{table}

\begin{figure}[!tbh]
    \centering
    \includegraphics[width=0.48\linewidth]{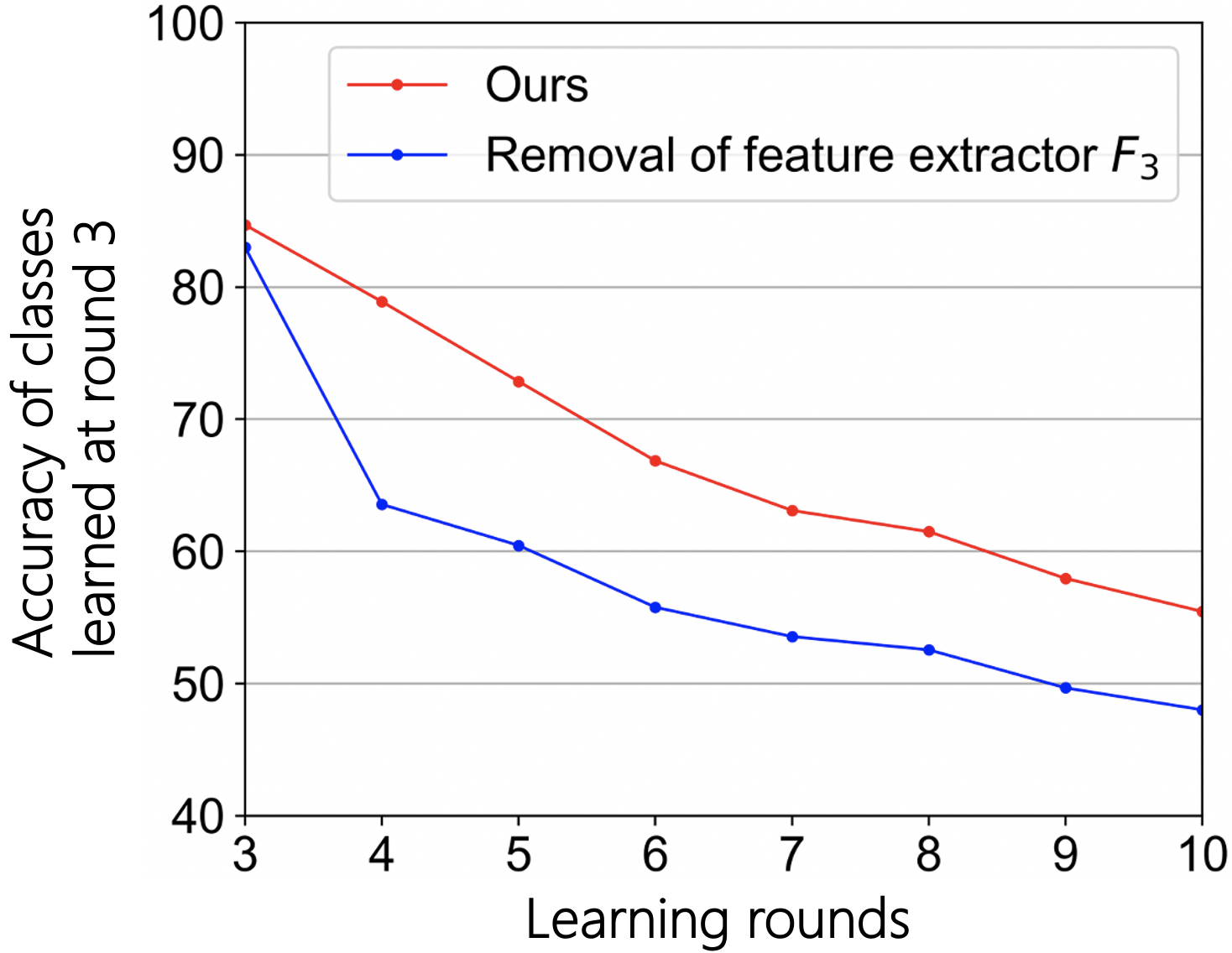} 
    \includegraphics[width=0.48\linewidth]{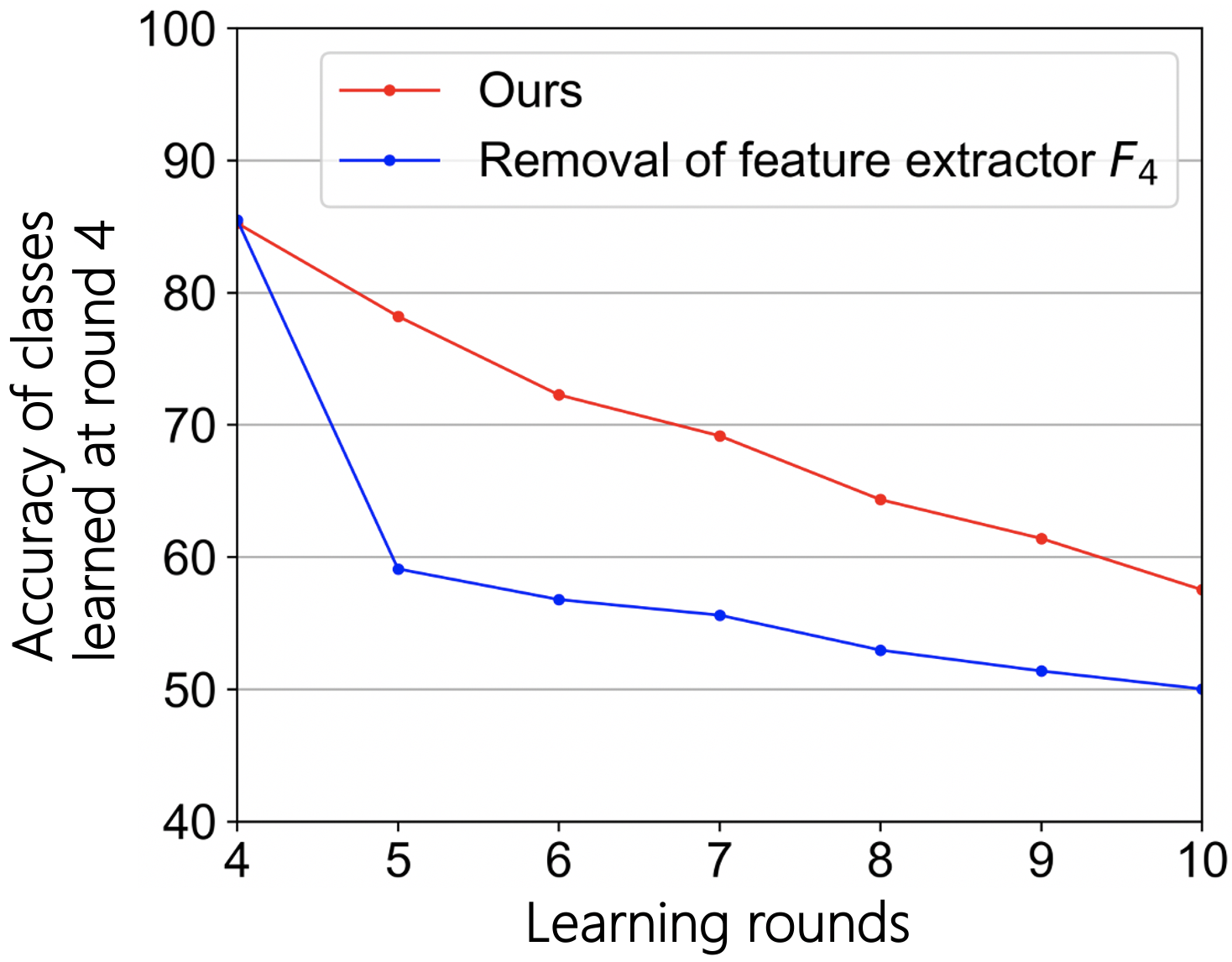}
\caption{Classification of the set of old classes after removing the associated feature extractor, over rounds of continual learning.}
\label{fig:ablation_fusion_wo_fn}
\end{figure}

\begin{figure}[!tbh]
    \centering
    \includegraphics[width=0.48\linewidth]{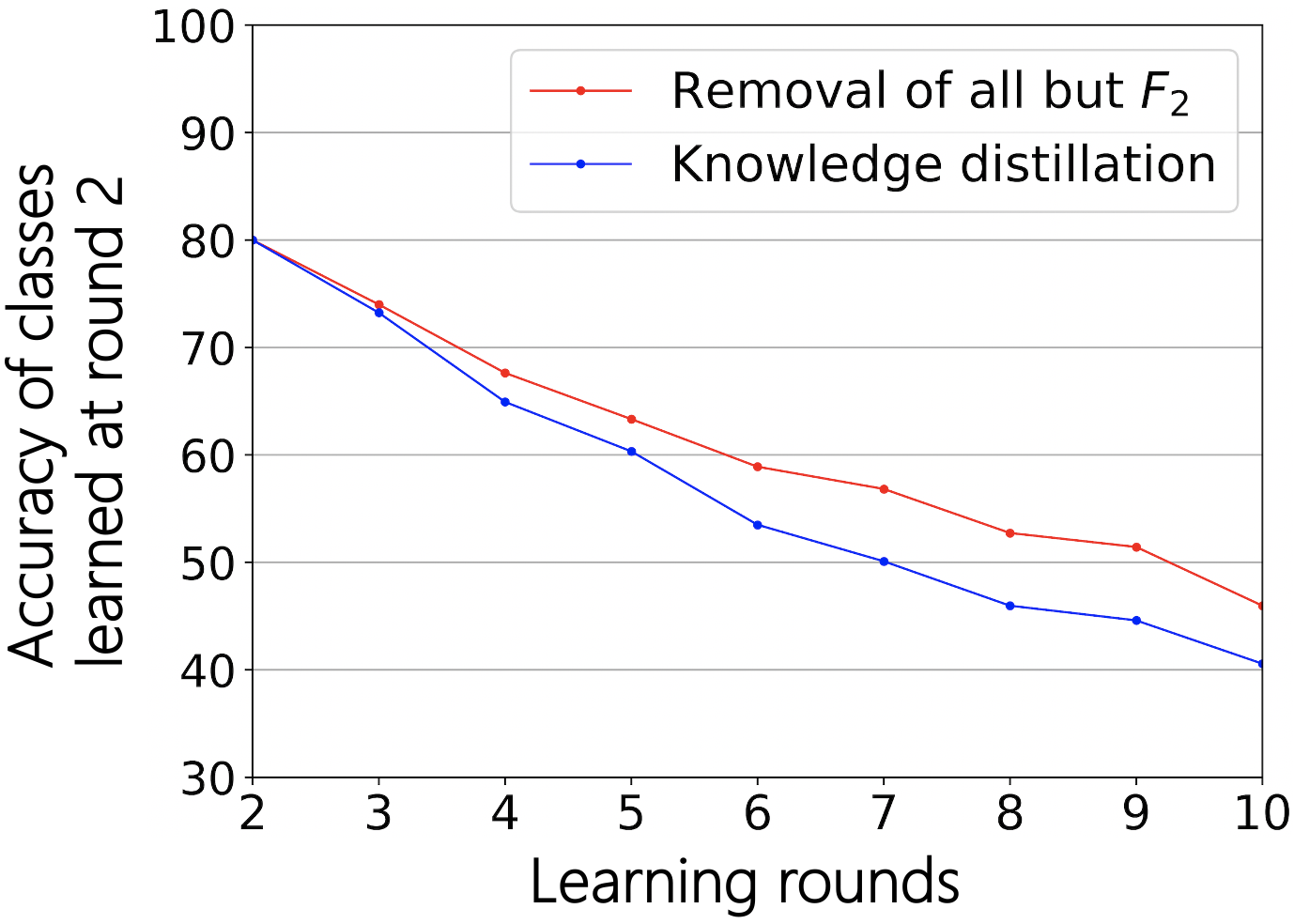} 
    \includegraphics[width=0.48\linewidth]{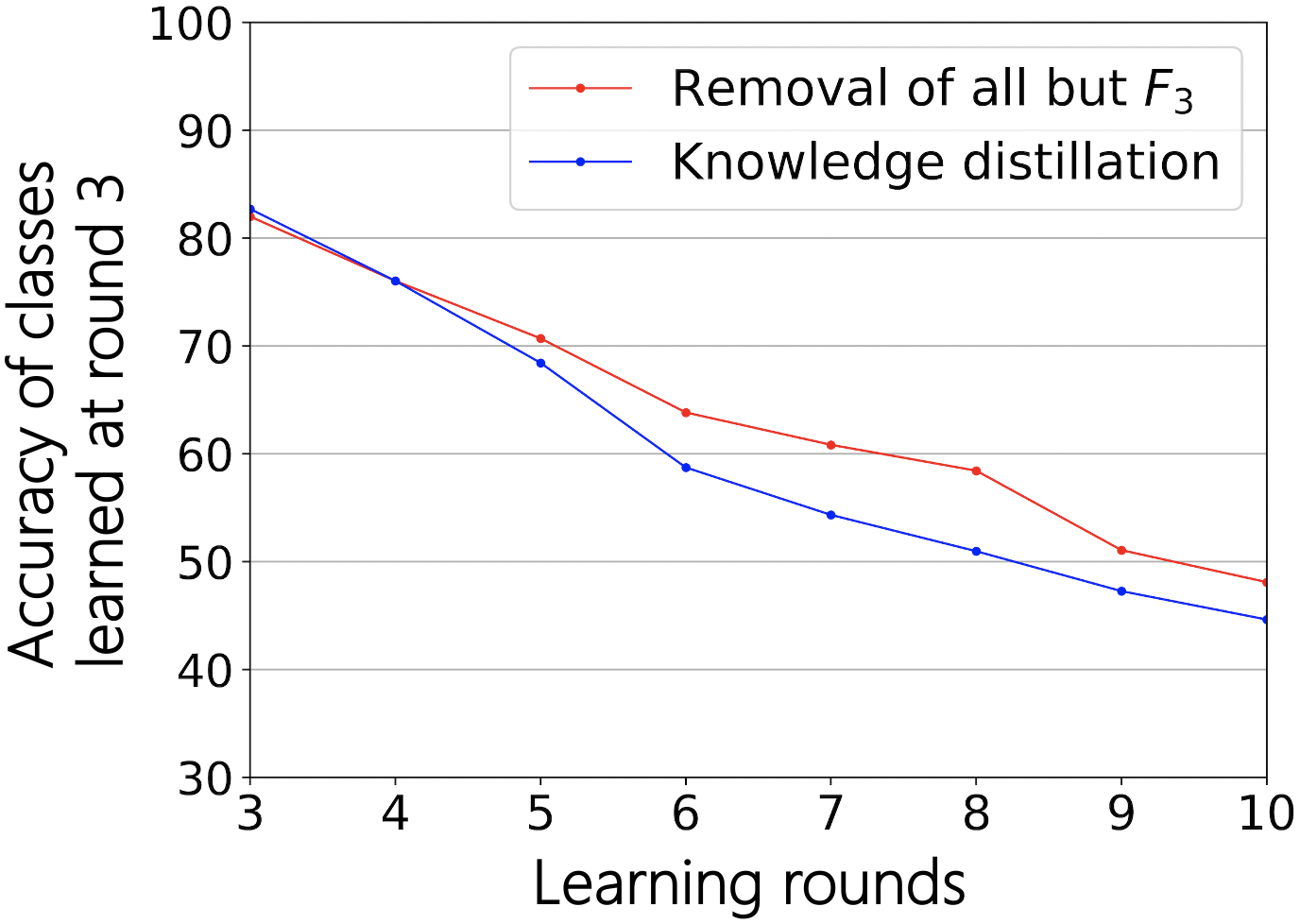}
\caption{Classification of the set of old classes after removing all but  the only associated feature extractor.}
\label{fig:ablation_fusion_w_fn}
\end{figure}

To further investigate the effect of previously learned feature extractors, two more experiments were performed, one with the removal of one specific previous feature extractor (Figure~\ref{fig:ablation_fusion_wo_fn}), and the other with the removal of all but one specific previous feature extractors (Figure~\ref{fig:ablation_fusion_w_fn}). 
As shown in Figure~\ref{fig:ablation_fusion_wo_fn}, after removing the feature extractor learned at round 3 (Left) or round 4 (Right), the knowledge of those classes learned at that round was less preserved (blue) compared to the inclusion of all feature extractors in the classifier (red). It suggests that the inclusion of a specific previous feature extractor does help preserve the knowledge of classes learned at that round. Consistently, when removing most previous feature extractors and just including one specific feature extractor, the classifier did perform relatively better on the old classes associated with the included feature extractor (Figure~\ref{fig:ablation_fusion_w_fn}, red) compared to the classifier without including any previous feature extractor (Figure~\ref{fig:ablation_fusion_w_fn}, blue), further confirming the effect of including any specific previous feature extractor.

\begin{figure}[!tbh]
    \centering
    \includegraphics[width=0.48\linewidth]{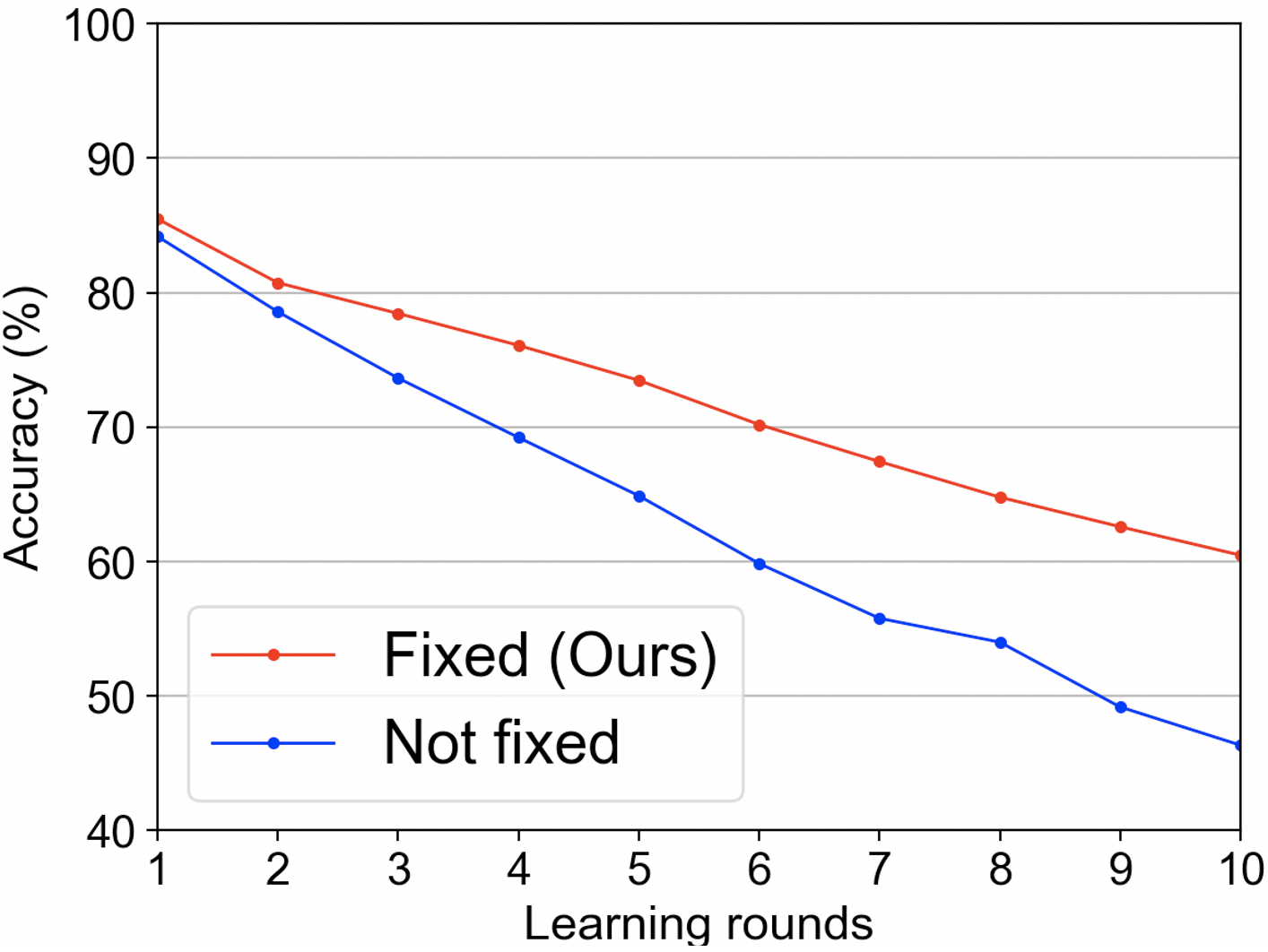}
    \includegraphics[width=0.48\linewidth]{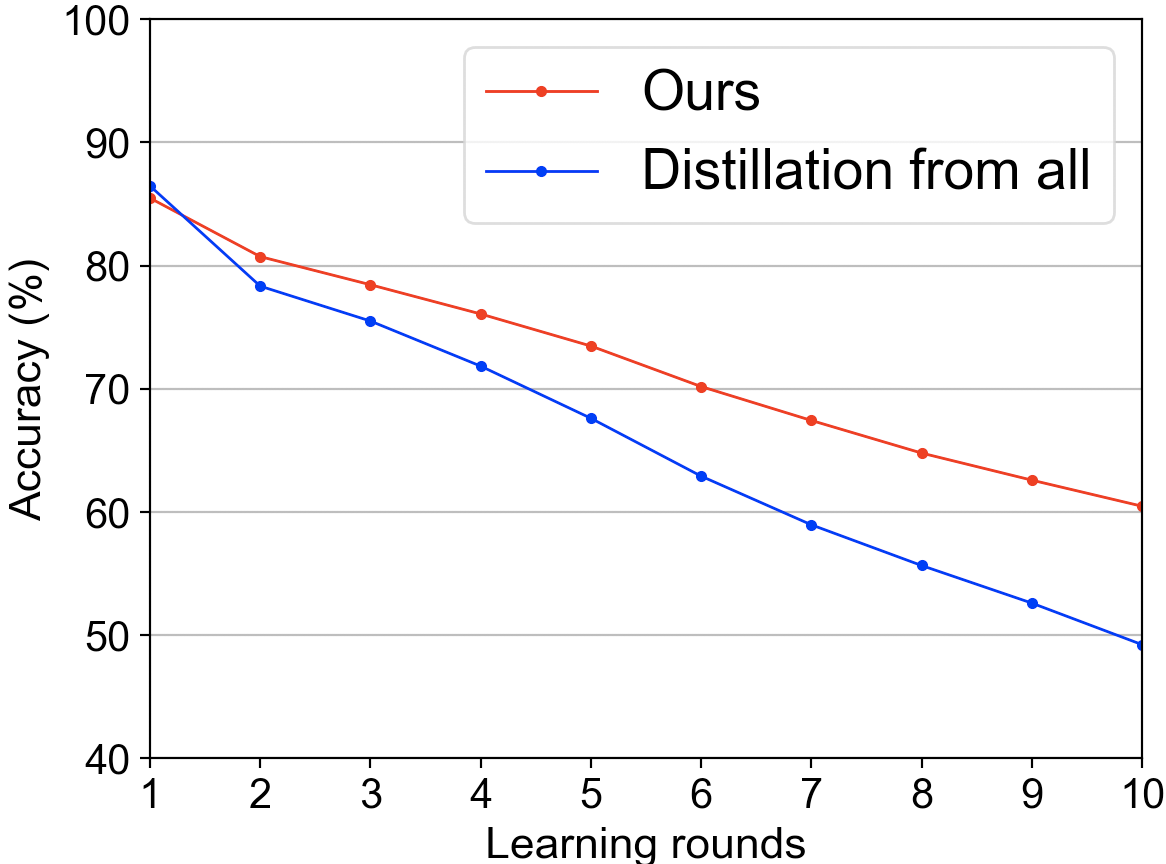}
    \caption{Necessity of fixing (Left) and keeping (Right) old feature extractors.}
    \label{fig:nofix_onefe}
\end{figure}

In addition, one may wonder whether the performance improvement comes from the increased number of model parameters contributed by these multiple feature extractors. Detailed investigation shows that, for each feature extractor, about 50$\sim$60\% kernels  and over $70\%$ of parameters were pruned (note that the pruned kernels at one layer also reduce kernel channels at the next layer). With ResNet18 as the backbone, the model size is respectively 0.28, 0.56, 0.83, 1.1, 1.3, 1.5, 1.7, 2.0, 2.2, and 2.5 times the size of the original ResNet18 model over 10 learning rounds, showing that the size-increasing speed is relatively low. Furthermore, Figure~\ref{fig:nofix_onefe} (Left) shows that, when allowing the old feature extractors to be fine-tuned during continual learning, the updated new classifier (blue curve) at each round performs much worse than that from the proposed method (red curve), further supporting that the performance improvement by the proposed method does not come from the increased model parameters but from the preserved old knowledge in the fixed old feature extractors. 

In contrast to the inclusion of all previously learned feature extractors into the classifier, one may wonder whether it is enough to directly distill knowledge from all the stored previous extractors to a single CNN classifier (i.e., with a single feature extractor). For example, to train a new single classifier $C_5$ which has 50 classes' output at round 5, the first $10 k$ logit (i.e., pre-softmax) outputs from $C_5$ and the whole logits from the $k$-th old classifier $C_k$ are used to form the $k$-th distillation loss term ($k=1,2,3,4$), and all such terms are averaged and added to the cross-entropy loss for $C_5$ training.  It is similar to End2End except that multiple (vs. single) distillation terms are used. 
Figure~\ref{fig:nofix_onefe} (Right) shows that such simple distillation strategy (blue curve) does not work well compared to the proposed fusion strategy, again supporting the effect of fusion of previous extractors on continual learning.

\section{Conclusion}
In this study, a simple yet effective continual learning framework was proposed by fusing all previously and currently learned feature extractors. The fusion of previous feature extractors helps reduce the forgetting of old knowledge particularly learned earlier. Comprehensive evaluations on multiple natural image classification tasks demonstrate  state-of-the-art continual learning performance by the proposed approach. Future work includes the application of the proposed approach to more challenging scenarios, e.g., continual learning of fine-grained classes and few-shot classes.


%



\section*{Acknowledgment}
This work is supported by the National Natural Science Foundation of China (grant No. 62071502, U1811461) and the Guangdong Key Research and Development Program (grant No. 2020B1111190001).

\ifCLASSOPTIONcaptionsoff
  \newpage
\fi



%

{\small
\bibliographystyle{IEEEtran}
\bibliography{IEEEfull.bib}
}




%








\end{document}